\documentclass[10pt,twocolumn,letterpaper]{article}

\usepackage[pagenumbers]{cvpr} 

\usepackage{xcolor}

\definecolor{cvprblue}{rgb}{0.21,0.49,0.74}
\usepackage[pagebackref,breaklinks,colorlinks,allcolors=cvprblue]{hyperref}
\usepackage[most]{tcolorbox}
\usepackage{caption}

\usepackage{amssymb}
\usepackage{pifont}

\usepackage{stfloats}

\tcbuselibrary{skins}

\newtcolorbox[auto counter, number within=section]{mybox}[2][]{
  colback=gray!5!white,
  colframe=gray!40!black,
  title={\textbf{Template~\thetcbcounter}: #2},
  fonttitle=\bfseries,
  #1
}

\def\confName{CVPR}
\def\confYear{2026}

\title{What Happens Next? Next Scene Prediction with a Unified Video Model}

\author{
\textbf{Xinjie Li}$^{1}$\thanks{Work done during an internship at Amazon.},
\textbf{Zhimin Chen}$^{2}$,
\textbf{Rui Zhao}$^{2}$\thanks{Corresponding author.},
\textbf{Florian Schiffers}$^{2}$,
\textbf{Zhenyu Liao}$^{2}$,
\textbf{Vimal Bhat}$^{2}$ \\
$^{1}$Pennsylvania State University, USA \quad
$^{2}$Amazon, USA \\
{\tt\small xql5497@psu.edu} \quad
{\tt\small \{zhiminch, zhaori, floschi, zyliao, vimalb\}@amazon.com}\\\\
\textbf{\url{https://nextsceneprediction.github.io/}}\\
}

\begin{document}

\twocolumn[{%
\renewcommand\twocolumn[1][]{#1}%
\maketitle
\vspace{-1.2cm}
\begin{center}
    \centering
    \captionsetup{type=figure}
    \includegraphics[width=0.95\linewidth]{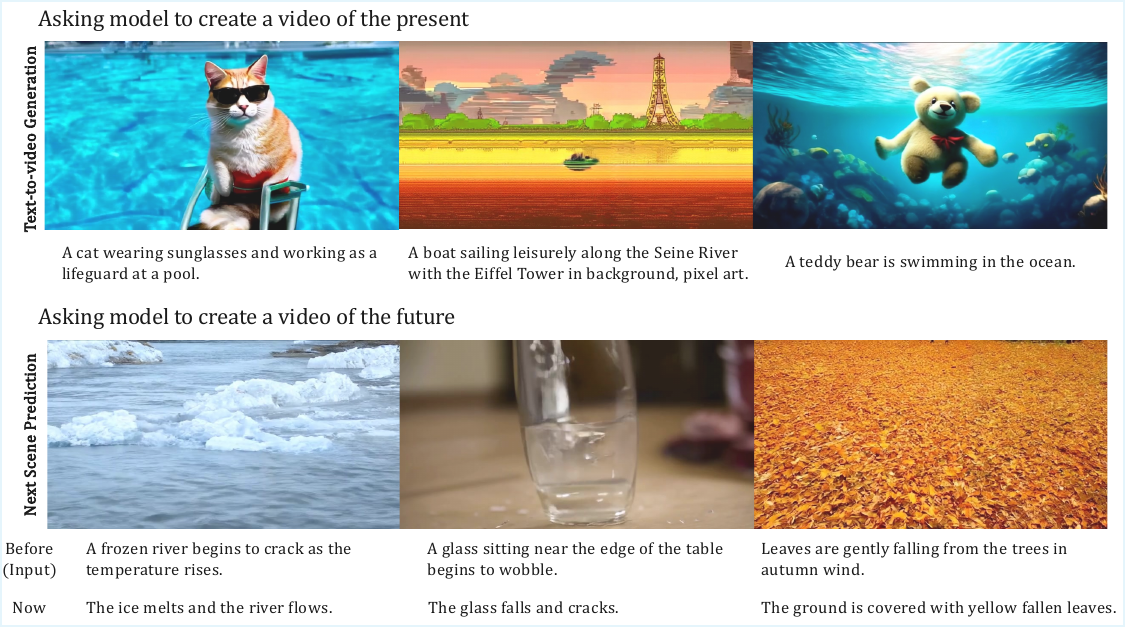}
    \vspace{-0.25cm}
    \captionof{figure}{Difference between text-to-video generation and our proposed next scene prediction task.
In text-to-video generation, the input caption and the generated video describe the same scene. In contrast, our next scene prediction task takes the preceding scene description as input and generates the subsequent scene video (note that the model only outputs the next-scene video; the texts in the figure serve for explanatory purposes). This task better aligns with the goal of a unified video model, as it requires stronger temporal reasoning and causal understanding across consecutive scenes.}
    \label{fig:teaser}

\end{center}%
}]

{
  \renewcommand{\thefootnote}{\fnsymbol{footnote}} 
  \footnotetext[1]{Work done during an internship at Amazon.}
  \footnotetext[2]{Corresponding author.}
}

\begin{abstract}

Recent unified models for joint understanding and generation have significantly advanced visual generation capabilities. However, their focus on conventional tasks like text-to-video generation has left the temporal reasoning potential of unified models largely underexplored. To address this gap, we introduce Next Scene Prediction (NSP), a new task that pushes unified video models toward temporal and causal reasoning. Unlike text-to-video generation, NSP requires predicting plausible futures from preceding context, demanding deeper understanding and reasoning. To tackle this task, we propose a unified framework combining Qwen-VL for comprehension and LTX for synthesis, bridged by a latent query embedding and a connector module. This model is trained in three stages on our newly curated, large-scale NSP dataset: text-to-video pre-training, supervised fine-tuning, and reinforcement learning (via GRPO) with our proposed causal consistency reward. Experiments demonstrate our model achieves state-of-the-art performance on our benchmark, advancing the capability of generalist multimodal systems to anticipate what happens next.

\end{abstract}

\section{Introduction}
\label{sec:intro}

The rapid advancement of generative AI has transformed creative workflows across art, design, and entertainment. Text-to-image and text-to-video models now enable users to visualize ideas and produce cinematic content directly from natural language prompts, democratizing creativity for artists and everyday users alike. Yet, despite their visual fidelity, current systems struggle with deeper aspects of creativity, such as spatial reasoning, semantic alignment, and context-aware editing. Real-world creation often requires iterative interaction, multimodal references, and fine-grained control, underscoring the need for models that not only generate visuals but also comprehend context and intent at a human-like level, such as designing a complex scene from multiple sketches, reference photos, and textual feedback.

In response to these challenges, the research community has increasingly turned toward unified multimodal models that integrate understanding and generation within a single framework~\cite{Zhang2025UnifiedMU, team2024chameleon,wu2025janus, Deng2025EmergingPI,Liao2025MogaoAO,Chen2025BLIP3oAF,Pan2025TransferBM,Inclusion2025MingLiteUniAI,Wu2025OpenUniAS}. Early progress~\cite{Deng2025EmergingPI,Liao2025MogaoAO,Chen2025BLIP3oAF,Pan2025TransferBM} has been primarily driven by image-based systems capable of both interpreting and producing visual content under language guidance. Recent advances have extended this paradigm toward video~\cite{luo2025univid,wei2025univideo,tan2025omni,Xiao2025HaploOmniUS}, aiming to unify video understanding and generation. Yet, most existing unified models remain focused on conventional tasks—such as text-to-image/video synthesis, image-to-video generation, and image/video editing—without fully exploiting their potential for temporal reasoning.

Motivated by this gap, we introduce a new task, \textbf{Next Scene Prediction (NSP)}, which aims to advance the temporal reasoning ability of unified multimodal models. Given a text prompt, \textit{``Please generate a video showing what happens after this scene: \textless preceding scene description\textgreater''}, the model is required to infer plausible future events and synthesize a coherent video that visualizes the causal continuation of the described scene. Unlike conventional text-to-video generation, which directly maps textual descriptions to visual content, NSP demands a deeper level of reasoning: the model must comprehend the temporal, causal, and semantic dynamics within the preceding scene before generating what follows.

This task introduces several key challenges. First, it requires \textit{temporal reasoning and causal inference}, as the model must predict how entities, actions, and environments evolve over time. Second, it demands \textit{deep semantic understanding} to maintain narrative consistency, emotion consistency, and contextual logic across scenes. Third, it requires \textit{visual plausibility}, ensuring that the generated future scenes are not only temporally consistent but also visually realistic. 
Together, these challenges make handling NSP task a critical capability toward building unified multimodal models that not only can depict what is described but also anticipate and create what comes next.

To tackle the challenges of NSP task, we design a \textit{unified video model} that integrates multimodal understanding and video synthesis within a single architecture. Specifically, we adopt {Qwen-VL}~\cite{bai2025qwen2} as the multimodal language model for scene understanding, and {LTX}~\cite{hacohen2024ltx} as the video diffusion model for efficient video generation, though other model architecture can be used under our generic framework. Furthermore, our framework also contains another two key components: a {latent query embedding}, which extracts and transfers high-level semantic representations from the multimodal language model, and a {connector module}, which bridges the understanding and generation models to enable end-to-end next scene prediction.

We employ a three-stage strategy to train the unified model: \textit{In the first stage}, We first pre-train the framework with the text-to-video task to align textual semantics with visual synthesis. \textit{In the second stage}, we then fine-tune the model on our curated NSP dataset to enhance its ability to reason about temporal and causal consistency. \textit{In the final stage}, we introduce \textit{causal consistency} reward measuring
how well the generated video preserves the causal and visual coherence with respect to the input preceding scene description.  Specifically, we employ the Group Relative Policy Optimization (GRPO)~\cite{shao2024deepseekmath} algorithm to maximize this causal consistency reward and further enhance the model’s temporal reasoning ability.

{Our contributions are summarized as follows:} 

\begin{enumerate}
\item We propose a novel \textit{NSP task} to explore the temporal reasoning potential of unified video models by requiring them to infer and generate a plausible future scene from preceding context. 

\item We design a \textit{unified framework} that integrates a multimodal language model and a video generation model, connected through a {latent query} embedding and a dedicated {connector} module to enable seamless information transfer. 

\item We develop a \textit{three-stage training} pipeline, consisting of (i) pre-training on text-to-video task, (ii) supervised fine-tuning on curated NSP dataset, and (iii) reinforcement learning with a \textit{causal consistency} reward. 

\item We curate a \textit{large-scale NSP dataset}, and experimental results on this dataset demonstrate the effectiveness of our approach.

\end{enumerate}

\section{Related Work}
\label{sec:related}

\subsection{Unified Multimodal Models}

\paragraph{Unified Image Models.}

Recent advances in unified image understanding and generation frameworks~\cite{Zhang2025UnifiedMU} exhibit several distinct paradigms. 

The first paradigm is the autoregressive encoder–decoder framework.
This line of work~\cite{Wang2025SelftokDV,Wang2025EndtoEndVT,team2024chameleon,wu2025janus,wang2024emu3} integrates autoregressive modeling for both textual and visual modalities. Input images are encoded via semantic or pixel-level encoders, while outputs are generated through pixel decoders or diffusion-based decoders—the latter offering superior visual fidelity. However, semantic encoders often lose fine-grained visual details, whereas pixel encoders lack high-level semantics, limiting their effectiveness in reasoning and editing tasks. 

The second paradigm is the unified hybrid architecture. State-of-the-art approaches~\cite{Deng2025EmergingPI,Liao2025MogaoAO,Tian2025UniGenET,Mo2025XFusionIN,cao2025hunyuanimage,xie2025show} unify autoregressive and diffusion mechanisms within a single transformer backbone. This design enables deeper synergy between multimodal understanding and generation, albeit at the cost of significantly higher computational demand. 

The third paradigm is the query/connector-based hybrid system.  
Recent methods~\cite{Chen2025BLIP3oAF,Pan2025TransferBM,Inclusion2025MingLiteUniAI,Wu2025OpenUniAS} bridge pre-trained understanding backbones with diffusion-based generators through learnable queries and/or connector modules. Despite that the connector can become an information bottleneck, these models enable efficient reuse of existing architectures. Our work follows this paradigm, allowing flexible integration of open-source video generators while remaining easier to train than fully unified end-to-end models.

\paragraph{Unified Video Models.}

Analogous to unified image models, unified video models can be broadly categorized into three paradigms.  
{Autoregressive frameworks}~\cite{Wu2024VILAUAU,Ge2024DivotDP,Jin2024VideoLaVITUV,tan2025omni} extend token-based autoregressive generation to the video domain. In contrast, {unified hybrid architectures}~\cite{Xiao2025HaploOmniUS,xie2025show} combine autoregressive modeling and diffusion-based synthesis within a single model to enhance multimodal video coherence. {Query/Connector-based hybrid systems}~\cite{luo2025univid,wei2025univideo} connect video understanding backbones with video generation modules via lightweight bridges. However, most prior works focus primarily on conventional text-to-video or video editing tasks, without exploring the temporal reasoning capabilities of unified video models.  
To address this gap, our work introduces a novel NSP task, which explicitly explores the reasoning potential of unified video models to infer and generate plausible future scenes.

\subsection{Reinforcement Learning for Visual Generation}

Reinforcement Learning from Human Feedback (RLHF) has emerged as a key approach for aligning generative visual models with human preferences. Early efforts directly fine-tune image and video generation models using scalar reward signals~\cite{Prabhudesai2023AligningTD, Clark2023DirectlyFD, Xu2023ImageRewardLA, prabhudesai2024video} or Reward Weighted Regression (RWR)~\cite{peng2019advantage, lee2023aligning, furuta2024improving}. Inspired by Proximal Policy Optimization (PPO)~\cite{schulman2017proximal}, policy gradient methods are later integrated into diffusion models~\cite{black2023training, fan2023reinforcement, gupta2025simple, miao2024training, zhao2025score}, demonstrating their ability to optimize complex perceptual objectives but suffering from high computational costs and sensitivity to hyperparameters. To improve efficiency, Direct Preference Optimization (DPO)~\cite{rafailov2023direct, wallace2024diffusion, dong2023raft, yang2024using, liang2024step, yuan2024self, liu2025videodpo, zhang2024onlinevpo} learns directly from human preference data via a supervised loss, bypassing explicit reward modeling. While DPO simplifies the training pipeline, it only optimizes relative preference rankings and is affected by confounding factors such as aesthetics or clarity. Recently, Group Relative Policy Optimization (GRPO)~\cite{shao2024deepseekmath} introduces group-wise advantage normalization to stabilize reinforcement learning optimization. In the visual generation domain, variants such as Flow-GRPO~\cite{liu2025flow} and Dance-GRPO~\cite{xue2025dancegrpo} extend GRPO to flow matching models by reformulating the underlying odrdinary differential equation (ODE) as  equivalent stochastic differential equation (SDE), enabling diverse sampling. MixGRPO~\cite{li2025mixgrpo} further improves training efficiency through a sliding-window mechanism. 
Building upon this line of research, we adapt GRPO to the NSP task, enabling optimization guided by causal consistency and improving the temporal coherence between the generated videos and the input preceding scene descriptions.

\section{Problem Statement}
\label{problem}
We first define NSP task as follows. Given a preceding scene description, the goal is to generate a temporally coherent and causally consistent video of subsequent scene. The model receives an instruction (i.e., NSP prompt):
\textit{``Please generate a video showing what happens after this scene: \textless preceding scene description\textgreater"}.
This task requires the model not only to synthesize visually consistent content but also to infer plausible temporal progression beyond the preceding scene.

\begin{figure*}[t]
  \centering
   \includegraphics[width=0.75\linewidth]{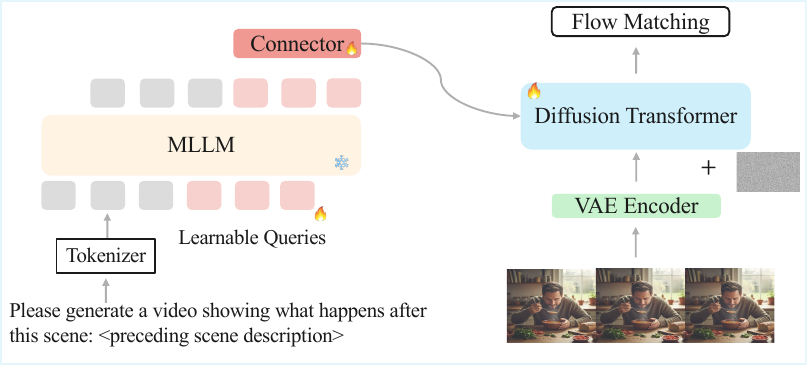}
   \vspace{-0.2cm}
   \caption{Framework illustration of our pre-training and supervised fine-tuning.
During pre-training, the input prompt is “Please generate a video based on the following caption: \textless caption\textgreater.” In the example shown, the preceding scene description is “The man was cooking a dish and just finished mixing the ingredients,” while the next scene in the video depicts the man tasting the dish to check its flavor.}
\vspace{-0.3cm}
\label{fig:framework1}
\end{figure*}

\section{A Unified Video Modeling Framework}
To tackle the NSP task, we propose a unified video model that integrates distinct components for multimodal understanding and video synthesis, illustrated in Fig.~\ref{fig:framework1}. The architecture consists of two primary interacting modules: (1) a powerful multimodal understanding backbone ({Frozen Qwen-VL}~\cite{bai2025qwen2}) enhanced with learnable query embeddings and (2) a fast, high-quality video generation engine ({LTX}~\cite{hacohen2024ltx}). These two modules are connected through a {connector} module that enables efficient cross-modal information flow and seamless cooperation between understanding and generation.
We now discuss each module in details.

\subsection{Multimodal Understanding Backbone}
We build our understanding backbone upon {Qwen-VL}, a large-scale pre-trained multimodal model designed to process and integrate information from text, image, and video modalities. In our setup, Qwen-VL is kept frozen to preserve its extensive pre-trained knowledge and multimodal reasoning capabilities. It receives two types of inputs: textual descriptions and a learnable query embedding.

\textbf{Textual Description.} The text input provide high-level semantic context and serve as conditioning signals that guide the video synthesis process, ensuring alignment between textual intent and visual generation.

\textbf{Learnable Query Embedding:} A learnable query embedding is proposed to dynamically interact with Qwen-VL’s attention layers, which give us the flexibility to adjust the understanding ability toward NSP. Essentially, the embedding is optimized during training to extract task-relevant features and adaptively focus on salient information for downstream video generation.

\subsection{Video Generation Engine}
\label{ltx}
Our generation component is built upon {LTX}, an efficient video generation model that operates in a compact latent space. Specifically, LTX utilizes a frozen Variational Autoencoder (VAE) to encode dense video frames into semantically rich latent representations and decode them back to high fidelity pixel space. On top of this latent space, a {Diffusion Transformer (DiT)} is employed as the denoising backbone, modeling the temporal and spatial dynamics of videos through diffusion processes.
It receives conditioning signals from the Qwen-VL backbone through a customized connector module (details in Sec.~\ref{sec-connector}), ensuring that video synthesis is semantically guided by the multimodal understanding process.

\subsection{Connector Module}
\label{connector}
\label{sec-connector}
We introduce a {connector} module that establishes the critical link between Frozen Qwen-VL and LTX. It transforms the high dimensional, semantically rich representations from Qwen-VL into a compact and properly scaled representations that LTX can effectively utilize. This facilitates coherent, contextually accurate video synthesis aligned with textual semantics.

Architecturally, the connector consists of two linear layers with a {SiLU}~\cite{Elfwing2017SigmoidWeightedLU} activation layer in between, followed by an RMSNorm~\cite{Zhang2019RootMS} layer and a learnable scale factor. This design helps stabilize training since the variance of Qwen-VL output embeddings are several orders of magnitude larger than that of the original LTX text encoder. The light-weighted architecture also enables efficient adaptation between the understanding and generation modules.

\section{Training Strategies}
\subsection{Preliminaries}
We first provide necessary background on the flow matching and GRPO, which is the basis for our unified model training.
\paragraph{Flow Matching. }
\label{fm}
Let $x_0 \in X_0$ denote a real data sample and $x_1 \in X_1$ a sample from a known prior distribution (e.g., Gaussian noise). Flow matching models~\citep{esser2024scaling} construct a continuous path between data and noise through linear interpolation:
\begin{equation}
\label{interpolation}
x_t = (1-t)x_0 + t x_1, \quad t \in [0,1].
\end{equation}
Differentiating with respect to $t$ yields the target velocity field:
\begin{equation}
\frac{dx_t}{dt} = x_1 - x_0.
\end{equation}
The objective is to learn a parameterized velocity field $v_{\theta}(x_t, t)$ that approximates this ground-truth flow by minimizing the following loss~\citep{lipman2023flow}:
\begin{equation}
\mathbb{L}(\theta) = \mathbb{E}_{t, x_0, x_1}\big[ \| (x_1 - x_0) - v_\theta(x_t, t) \|^2 \big].
\end{equation}
Unlike diffusion models~\citep{ho2020denoising}, which estimate a score function or noise term, flow matching directly learns the velocity of the underlying probability flow ODE. This provides a more explicit and stable parameterization of the generative process, often resulting in faster convergence and improved training stability in practice.

\paragraph{Group Relative Policy Optimization (GRPO). }
\label{grpo}

To stably optimize generative models under reinforcement feedback, GRPO~\citep{shao2024deepseekmath} extends PPO~\citep{schulman2017proximal} by introducing group-wise advantage normalization. 
Given a prompt condition $\boldsymbol{c}$, the flow matching model generates a set of samples $\{\boldsymbol{x}_0^i\}_{i=1}^{G}$ from noise inputs $\{\boldsymbol{x}_1^i\}_{i=1}^{G}$ via the velocity field $\boldsymbol{v}_{\theta}(\boldsymbol{x}_t, \boldsymbol{c}, t)$:
\begin{equation}
\text{d}\boldsymbol{x}_t = \boldsymbol{v}_{\theta}(\boldsymbol{x}_t, \boldsymbol{c}, t)\text{d}t, \quad t \in [0,1].
\end{equation}

The reward model assigns a scalar score $r(\boldsymbol{x}_0^i, \boldsymbol{c})$ to each generated sample, reflecting its alignment with human preferences or perceptual objectives. 
Within each group, the normalized advantage for sample $i$ is computed as:
\begin{equation}
\hat{A}_t^i = 
\frac{r(\boldsymbol{x}_0^i, \boldsymbol{c}) - \operatorname{mean}\!\left(\{r(\boldsymbol{x}_0^j, \boldsymbol{c})\}_{j=1}^{G}\right)}
{\operatorname{std}\!\left(\{r(\boldsymbol{x}_0^j, \boldsymbol{c})\}_{j=1}^{G}\right)}.
\end{equation}
This intra-group normalization reduces reward scale variance and stabilizes optimization by comparing relative sample quality within each group.

The policy model is then updated to maximize the clipped surrogate objective:
\begin{equation}
\label{eq:grpo_objective}
\begin{aligned}
\mathcal{J}(\theta)
= 
\mathbb{E}_{\{\boldsymbol{x}^i\}_{i=1}^G \sim \pi_{\theta_{\text{old}}}(\cdot|\boldsymbol{c})}
\frac{1}{G} \sum_{i=1}^{G} \frac{1}{T}\sum_{t=0}^{T-1}
\min\!\Big(
&\rho_t^i(\theta)\hat{A}_t^i, \\
\operatorname{clip}\!\big(\rho_t^i(\theta), 1-\varepsilon, 1+\varepsilon\big)\hat{A}_t^i
\Big),
\end{aligned}
\end{equation}
where $
\rho_t^i(\theta) = 
\frac{p_{\theta}(\boldsymbol{x}_{t-1}^i|\boldsymbol{x}_t^i, \boldsymbol{c})}
     {p_{\theta_{\text{old}}}(\boldsymbol{x}_{t-1}^i|\boldsymbol{x}_t^i, \boldsymbol{c})},
$ is the likelihood ratio between updated policy $p_{\theta}$ and old policy $p_{\theta_{\text{old}}}$ and $\varepsilon$ is a constant controlling the clipping range. 

To apply GRPO to flow matching models, we adopt \cite{liu2025flow,xue2025dancegrpo} to replace the deterministic ODE with its stochastic counterpart:
\begin{equation}
\label{eq:grpo_sde}
\begin{aligned}
\text{d}\boldsymbol{x}_t 
= \Big[
&\boldsymbol{v}_{\theta}(\boldsymbol{x}_t, \boldsymbol{c}, t)
+ \frac{\sigma_t^2}{2t}\big(\boldsymbol{x}_t + (1-t)\boldsymbol{v}_{\theta}(\boldsymbol{x}_t, \boldsymbol{c}, t)\big)
\Big]\text{d}t \\
&\quad + \sigma_t \text{d}\boldsymbol{w},
\end{aligned}
\end{equation}
where $\sigma_t$ introduces stochasticity during sampling and $\text{d}\boldsymbol{w}$ denotes Wiener process increments. 
This reformulation enables diverse trajectory exploration and smooth gradient updates, effectively combining the sample diversity of SDEs with the training stability of ODE-based flow matching.

\begin{figure}[t]
  \centering
   \includegraphics[width=1\linewidth]{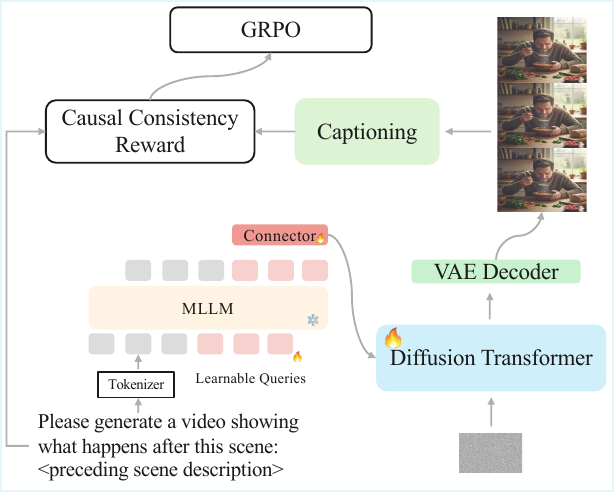}
   \vspace{-0.5cm}
   \caption{Framework illustration of our reinforcement learning stage. Note that this stage corresponds to the sampling phase in standard flow matching model training.}
\vspace{-0.5cm}
\label{fig:framework2}
\end{figure}

\subsection{Training}
We train our model in three stages: first, we pre-train it on the text-to-video task to establish a strong backbone; next, we perform supervised fine-tuning on our curated NSP dataset to adapt the model to the new task; and finally, we apply reinforcement learning to further enhance its temporal reasoning ability.

\paragraph{Pre-training.}
We first pre-train our proposed unified model with the text-to-video generation task, following a {curriculum learning strategy} that progressively aligns semantics across modalities and adapts the model to video generation. The process consists of three stages:

\begin{itemize}
    \item {Stage I – Text-to-Image training:} The connector is first pre-trained using large-scale text-to-image generation datasets to align textual semantics with the generation model’s latent space, thereby expanding its concept coverage. Note that only the connector is trained here.
    \item {Stage II – Joint Image-Video training:} We jointly train the connector and diffusion transformer (DiT) on a mixture of text-to-image and text-to-video generation datasets, gradually adapting the model from static image synthesis to dynamic video modeling.
    \item {Stage III – Text-to-Video training:} The connector and DiT are jointly trained on text-to-video generation datasets to enable fine-grained video modeling.
\end{itemize}

Throughout all stages, we adopt lightweight versions of generation model (Sec.~\ref{ltx}) and connector (Sec.~\ref{connector}) to enhance training efficiency without compromising representation power.
During this text-to-video pre-training stage, the input instruction is \textit{``Please generate a video based on the following caption: \textless caption\textgreater ''}. The model is trained with flow matching as discussed in Sec.~\ref{fm}.

\paragraph{Supervised Fine-tuning (SFT).}
We then fine-tune the text-to-video generation model using our curated NSP dataset. The dataset contains pairs of preceding scene descriptions and corresponding next-scene video clips, allowing the model to learn progression patterns and causal consistency. Details of NSP dataset are introduced in Sec.~\ref{data}.
During this supervised fine-tuning stage, the NSP task prompt (Sec.~\ref{problem}) is used as the input instruction. The model is also trained with flow matching as discussed in Sec.~\ref{fm}.

\paragraph{Reinforcement Learning with GRPO.}

To further improve temporal reasoning, we apply GRPO with feedback from an existing multimodal LLM (i.e., judge model). The process is shown in Fig.~\ref{fig:framework2}. During this reinforcement learning stage, the NSP task prompt (Sec.~\ref{problem}) is used as the input instruction, and the GRPO algorithm (Sec.~\ref{grpo}) is used for training. A novel reward function is introduced as follows.

\textbf{Causal Consistency Reward. }
To determine the reward for a generated video $v$, we first obtain its textual caption $q$ using the judge model prompted with:
\textit{``Describe this video in detail. Focus on the actions, objects, people, and what is happening in the scene. Provide a coherent description of the video content."}
We then use the judge model as a reward function to assess the semantic and causal consistency between the generated video caption $q$ and the input preceding scene description $s$, with the prompt template defined in Template~\ref{metric-full} (supplementary material).
Finally, the response from the judge model is parsed into a binary reward function:
\begin{align}
r(s, q) =
\begin{cases}
1, & \text{if verdict = ``Pass''} \\
0, & \text{if verdict = ``Fail''}.
\end{cases}
\end{align}

The model is then optimized via GRPO to maximize the expected reward. 
This reinforcement learning stage encourages the model to produce temporally coherent continuations that are semantically novel and causally plausible.

\section{Datasets Curation}
\label{data}

\paragraph{Pre-training Datasets.}

For pre-training, we first utilize the BLIP3o Long Caption dataset~\cite{Chen2025BLIP3oAF} and split it into two parts: the first part is used for Stage~1 text-to-image training, while the second part is used for Stage~2 image–video mixed training. 
In Stage~2, we also incorporate the VidGen~\cite{tan2024vidgen}, OpenVid~\cite{nan2024openvid}, and OpenS2V~\cite{yuan2025opens2v} datasets for text-to-video generation. 
Finally, in Stage~3, we employ the OpenHumanVid~\cite{Li2024OpenHumanVidAL} dataset, which is human-centric and complements the limited coverage of human-related scenarios in previous stages. 
The data filtering criteria for quality assurance and a detailed summary of all datasets are provided in
the supplementary material.

\paragraph{SFT and RL Datasets.}
We construct a new dataset for our NSP task, derived from existing text-to-video generation datasets (OpenS2V~\cite{yuan2025opens2v}, OpenVid~\cite{nan2024openvid}). 
To prevent information leakage, this dataset is curated independently from those used in pre-training. 
Existing text-to-video generation datasets consist of caption–video pairs. In our curation process, we treat such pairs as the subsequent scene video and its description. We create a preceding scene description that logically leads to the subsequent scene. 

To achieve this, we employ an LLM prompted by commonsense-driven templates to generate the preceding scene description that is causally related yet non-redundant to the subsequent scene. We further verify the generated preceding scene description using an LLM-based filtering process that checks for both redundancy and causal consistency compared to the subsequent scene description. Samples failing the verification after three iterations are discarded. Prompts for preceding scene description generation and verification are provided in the supplementary material. To ensure quality, we also filter video in the same way as pre-training dataset.

In total, the resulting NSP dataset contains 0.97M samples for SFT and 8K samples for RL. Each sample contains a preceding scene description and a next-scene video. Data samples are provided in the supplementary material.

\section{Experiments}
\label{sec:exp}

\subsection{Evaluation}

For text-to-video generation, we employ \textit{VBench}~\cite{huang2023vbench} to evaluate both quality and semantic alignment scores.
For the NSP task, we measure \textit{causal consistency}, defined as the proportion of samples labeled as “Pass” by the judge model, according to the causal-consistency criteria outlined in the reward prompt (Template~\ref{metric-full} in supplementary material).

\subsection{Main Results}

\paragraph{Text-to-Video Generation.}

Although our primary focus is on the NSP task, we still evaluate our pre-trained text-to-video model against the original LTX baseline to verify that our modifications do not degrade generation quality. As shown in
Table~\ref{tab:nsp-sota}, our model produces videos with improved visual quality and semantic alignment, demonstrating the superiority of replacing the original text encoder and introducing our unified architecture. Visual comparisons are shown in supplementary material. During evaluation, a CFG scale of 3.0 is applied. For fair comparison, we generate 65 frames at a resolution of 832$\times$480 without negative guidance, using 50 sampling steps.

\begin{table}
  \caption{Comparison between our pre-trained text-to-video model and the original LTX model on the VBench benchmark.
}
\vspace{-0.3cm}
  \label{tab:nsp-sota}
  \centering
  \begin{tabular}{@{}lccc@{}}
    \toprule
    Models & Quality& Semantic& Total \\
    \midrule
    LTX & 0.7799& 0.6738& 0.7586 \\
    Ours & 0.8051 & 0.6945& 0.7830\\
    \bottomrule
  \end{tabular}
  \vspace{-0.2cm}
\end{table}

\paragraph{Next Scene Prediction.}

\begin{figure*}[t]
  \centering
\includegraphics[width=0.9\linewidth]{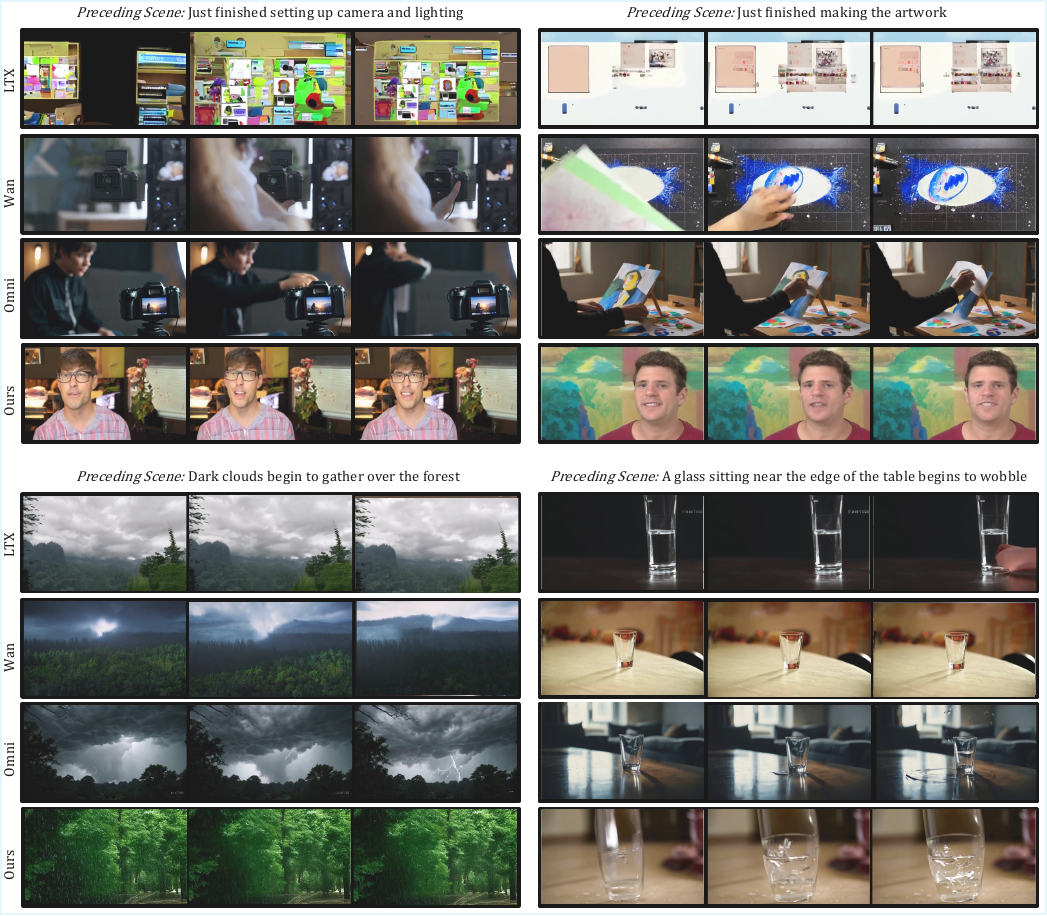}
   \vspace{-0.2cm}
   \caption{
Qualitative comparisons between different methods on our NSP task. Previous methods tend to generate videos that simply follow the given description, whereas our approach can infer and synthesize the subsequent scene. In the first row, the left video shows our model generating a man in an interview, while the right video depicts a man introducing his artwork. In the second row, the left video shows rainfall, while the right video depicts a glass cracking.
Video results are attached in the supplementary material.}
   \vspace{-0.2cm}
   \label{fig:sota}
\end{figure*}

We curate an additional 1K samples that are not used and distinct from any training data for evaluation. In this setting, we primarily compare our model with text-to-video generation models, including the original LTX model~\cite{hacohen2024ltx}, the Wan 2.1 1.3B model~\cite{Wang2025WanOA}, and the open-source unified video model Omni-Video~\cite{tan2025omni}. As shown in Table~\ref{tab:t2v-sota}, our model outperforms previous methods by a significant margin. As illustrated in Fig.~\ref{fig:sota}, previous methods tend to strictly follow the input description, whereas our model can infer and generate the subsequent scene, demonstrating stronger causal reasoning and temporal understanding. 

\begin{figure*}
  \centering
   \includegraphics[width=0.85\linewidth]{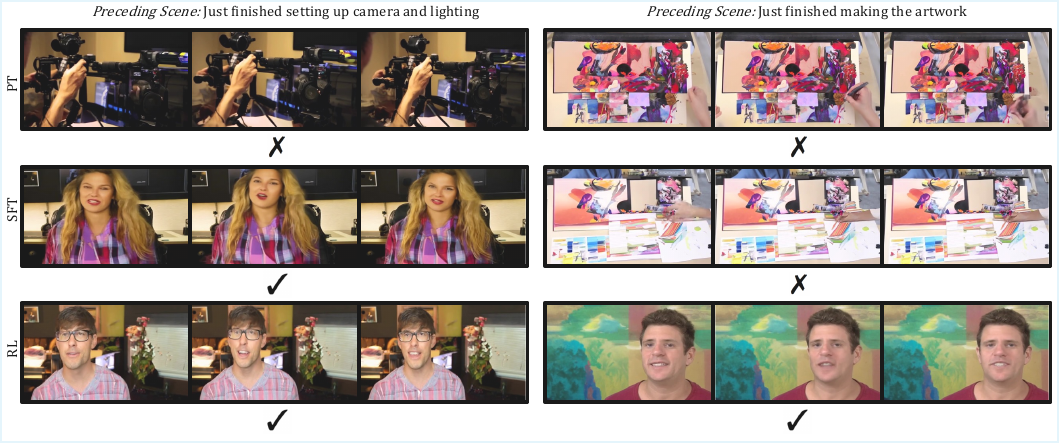}
   \vspace{-0.2cm}
   \caption{Qualitative comparisons between different stages on our NSP task. \ding{55} indicates that the generated video does not depict the next scene, while \ding{51} denotes correct next-scene video generation. Video results are attached in the supplementary material.}
   \label{fig:ablation}
   \vspace{-0.2cm}
\end{figure*}

\begin{table}
  \caption{Causal consistency comparison on our NSP test dataset.
}
\vspace{-0.2cm}
  \label{tab:t2v-sota}
  \centering
  \begin{tabular}{@{}lc@{}}
    \toprule
    Methods & Causal Consistency \\
    \midrule
    LTX &  0.23 \\
    Wan & 0.25 \\
    Omni-Video & 0.46 \\
    Ours & 0.73 \\
    \bottomrule
  \end{tabular}
  \vspace{-0.5cm}
\end{table}

\subsection{Ablation Analysis}

\paragraph{Different Training Stages for Pre-training.}

As shown in Table~\ref{tab:t2v-abla}, the first stage, which uses only image data, establishes a solid foundation for subsequent training, achieving reasonable quality and semantic scores. Incorporating text-to-video datasets in stage 2 further enhances both visual quality and semantic alignment. In stage 3, incorporating human-centric video data brings further improvements, primarily compensating for the lack of human-related content in earlier stages.

\begin{table}
  \caption{Comparison across different training stages of pre-training on the VBench benchmark.
}
\vspace{-0.2cm}
  \label{tab:t2v-abla}
  \centering
  \begin{tabular}{@{}lccc@{}}
    \toprule
    Training Stage & Quality& Semantic& Total \\
    \midrule
    Stage I & 0.7824& 0.5494& 0.7358 \\
    Stage II & 0.8019 & 0.6861& 0.7788\\
    Stage III & 0.8051 & 0.6945& 0.7830\\
    \bottomrule
  \end{tabular}
  \vspace{-0.2cm}
\end{table}

\begin{table}
  \caption{Causal consistency comparison of different training stages for our proposed NSP task on the NSP test dataset.
}
\vspace{-0.2cm}
  \label{tab:nsp-abla}
  \centering
  \begin{tabular}{@{}lc@{}}
    \toprule
    Training Stage & Causal Consistency \\
    \midrule
    Pre-training & 0.54  \\
    SFT & 0.60 \\
    RL & 0.73 \\
    \bottomrule
  \end{tabular}
  \vspace{-0.2cm}
\end{table}

\paragraph{Different Stages for NSP.}

\begin{figure}
  \centering
\includegraphics[width=1\linewidth]{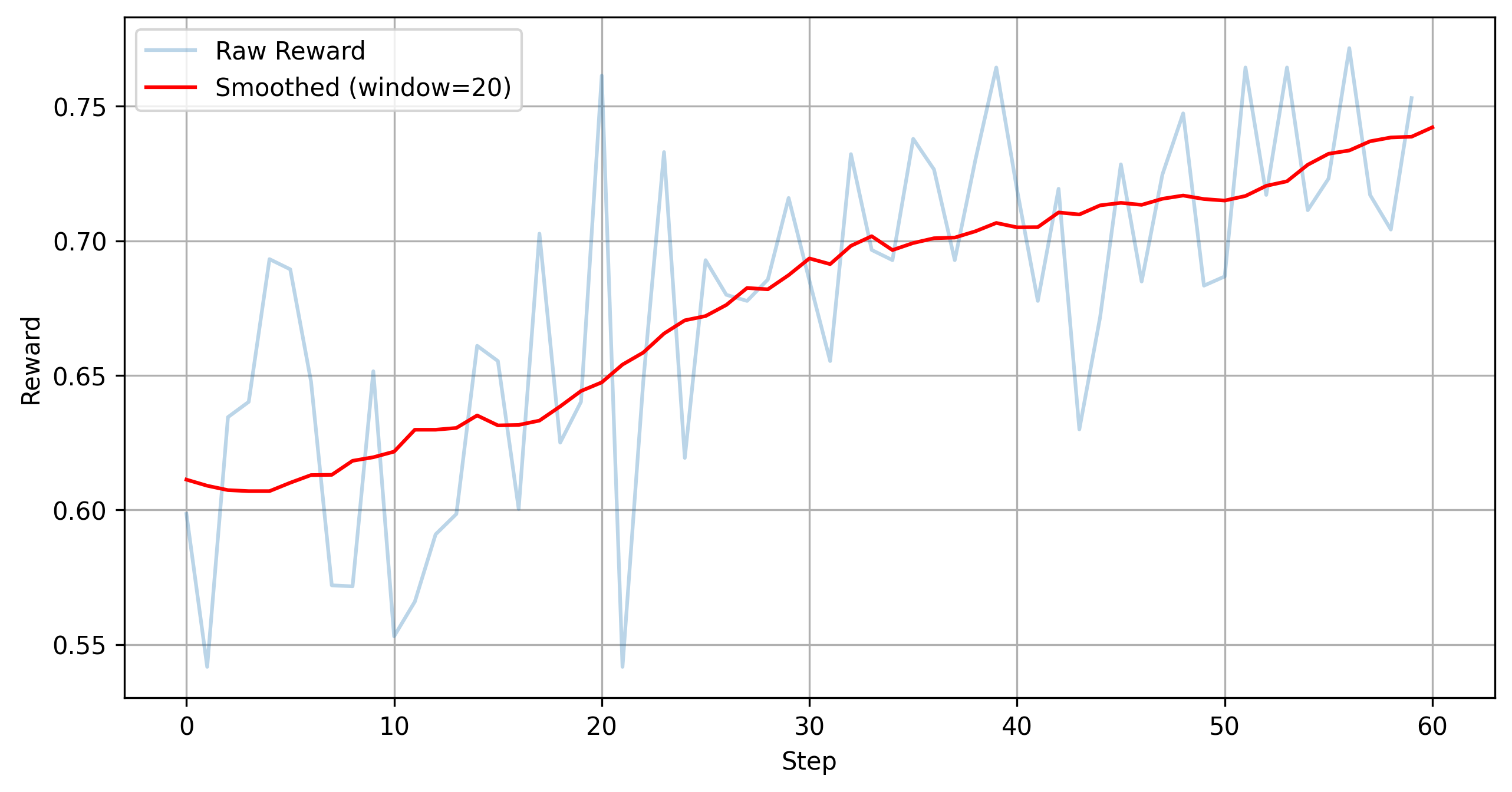}
\vspace{-0.7cm}
   \caption{We visualize the training curve of our proposed causal consistency reward, which initially exhibits fluctuations but soon stabilizes and steadily improves over time.}
   \label{fig:reward}
   \vspace{-0.5cm}
\end{figure}

As shown in Table~\ref{tab:nsp-abla}, both SFT and RL enhance causal consistency. As illustrated in Fig.~\ref{fig:ablation}, the pre-trained model mainly follows the input description, whereas supervised fine-tuning and reinforcement learning enable the model to generate more causally consistent and coherent next-scene videos. The reward curve during the RL stage is presented in Fig.~\ref{fig:reward}.

\section{Discussions}
\label{discussion}

\begin{enumerate}

    \item The proposed NSP task could potentially serve as an effective proxy for improving comprehension. By design, the NSP task compels the model to master temporal and causal dependencies, thereby coupling its understanding and generation faculties. Though the understanding module in our framework is not trained, this synergy warrants further investigation.

    \item We employ text-to-video pre-training for a robust foundation, but a domain mismatch exists: standard text-to-video synthesis relies on detailed captions, while our task uses concise preceding scene descriptions. This divergence may cause visual artifacts, a limitation that remains an open question for future research.
    
    \item Our work serves as a starting point for exploring causal temporal reasoning in unified video models. Future work can benefit from more rigorous evaluation metrics, larger model backbones, improved reward functions, and expanded datasets.
\end{enumerate}

\section{Conclusion}
\label{sec:conc}

We propose a novel NSP task to advance temporal and causal reasoning in unified video models. Our unified framework connects Qwen-VL and LTX (via a latent query embedding and connector) and is trained using a multi-stage training, i.e., pre-training, supervised fine-tuning, and reinforcement learning with our proposed causal consistency reward. We also propose a large-scale NSP dataset for training and evaluation. Experiments demonstrate clear gains in reasoning and coherence over baselines on our benchmark. We conclude that NSP offers a new path toward multimodal systems that can understand, reason, and generate future-consistent visual content.

{
    \small
    \bibliographystyle{ieeenat_fullname}
    \bibliography{main}
}

\clearpage
\setcounter{page}{1}
\maketitlesupplementary

\section{Causal Consistency Reward}

Here, we present the prompt for causal consistency reward. Given a preceding scene description and a generated video caption, the prompt requires the judge model to evaluate whether the two are logically consistent. In addition, it requires verifying that the next scene does not excessively repeat the content of the preceding one.

\begin{mybox}[label=metric-full]{Causal Consistency Reward}
\footnotesize
\texttt{
You will be given two text descriptions: \\
(1) A preceding scene (context) \\
(2) A generated video description (caption) \\
Your task is to determine whether there is logical causal consistency between the scene and caption. Additionally, evaluate whether the caption is semantically redundant with the scene description---that is, whether it repeats most of the same people, actions, objects, or setting, rather than continuing with new content.\\
\textbf{Criteria:} \\
\textbf{Pass} if: \\
\quad -- The caption shows logical causal continuity from the scene description. \\
\quad -- The caption describes a clearly new moment with different actions, progression, or consequences. \\
\quad -- The overlap in content (people, actions, objects, setting) is low. \\
\textbf{Fail} if: \\
\quad -- The caption is disconnected or implausible in progression. \\
\quad -- The caption largely repeats the same content. \\
\quad -- The two descriptions refer to the same moment or situation. \\
\textbf{Input:} \\
Scene Description: \{scene\_description\} \\
Caption: \{caption\} \\
\textbf{Output:} \\
Verdict: Pass / Fail
}
\end{mybox}

\section{NSP Dataset Curation}

\subsection{Prompts}
Here, we present the prompts for generating preceding scene descriptions and for verification.

 \begin{tcolorbox}[colback=gray!4!white, colframe=gray!60!black, title={Preceding Description Prompt Template}, fonttitle=\bfseries]
\ttfamily
\small
Task: Predict the most likely prior event or activity using commonsense knowledge based on the observed video. \\
The video shows: [Current visual scene or activity] \\
Description: [What likely happened just before this moment] \\=
Guidelines: Use real-world causal reasoning. Describe an earlier event that {leads to} the current scene, without repeating existing actions or characters.\\
Example: \\
The video shows: a boy is tying his shoelaces.\\
Description: He was in a hurry to catch the school bus outside.
\end{tcolorbox}

\begin{tcolorbox}[colback=gray!4!white, colframe=gray!60!black, title={Verification Prompt Template}, fonttitle=\bfseries]
\ttfamily
\small
You will be given two text descriptions: \\
1. A preceding scene (context) \\
2. A generated video description (caption) \\
Your task: Determine whether the caption is semantically redundant with the scene description or if it provides new, causally consistent content. \\
Label as ``Pass'' if the caption describes a new moment with causal continuity; otherwise label as ``Fail''.
\end{tcolorbox}

\subsection{Data Filtering}
To ensure high-quality supervision, we apply the following filtering rules:
(1) retain video clips shorter than 5 seconds at 16 fps (maximum 81 frames); for longer clips, only the first 5 seconds are used;
(2) keep videos with a resolution of at least 832$\times$480 pixels;
(3) resize all training videos to 832$\times$480 during training. Each image in the image datasets is treated as a single-frame video with a resolution of 512$\times$512. While the current setup adopts fixed spatial resolutions, future extensions will explore adaptive resolution training.

\section{Text-to-video Pre-training}

\subsection{Dataset Statistics}

Detailed dataset statistics for the different training stages for text-to-video pre-training are presented in Table~\ref{tab:datasets}.

\begin{table*}[t]
\centering
\caption{Summary of datasets used in different training stages for text-to-video pre-training.}
{
\begin{tabular}{lccccc}
\toprule
\textbf{Dataset} &  \textbf{Quantity}   & \textbf{Stages}  \\
\midrule
BLIP-3o (Image)  & 27M  & 20M (Stage 1), 7M (Stage 2) \\
 VidGen  & 1.0M &  Stage 2 \\
 OpenVid & 0.31M &  Stage 2\\
OpenS2V & 1.33M &  Stage 2 \\
OpenHumanVid  & 10.8M &  Stage 3\\
\bottomrule
\end{tabular}
}
\label{tab:datasets}
\end{table*}

\subsection{Results}
\begin{figure*}[t]
  \centering
   \includegraphics[width=0.8\linewidth]{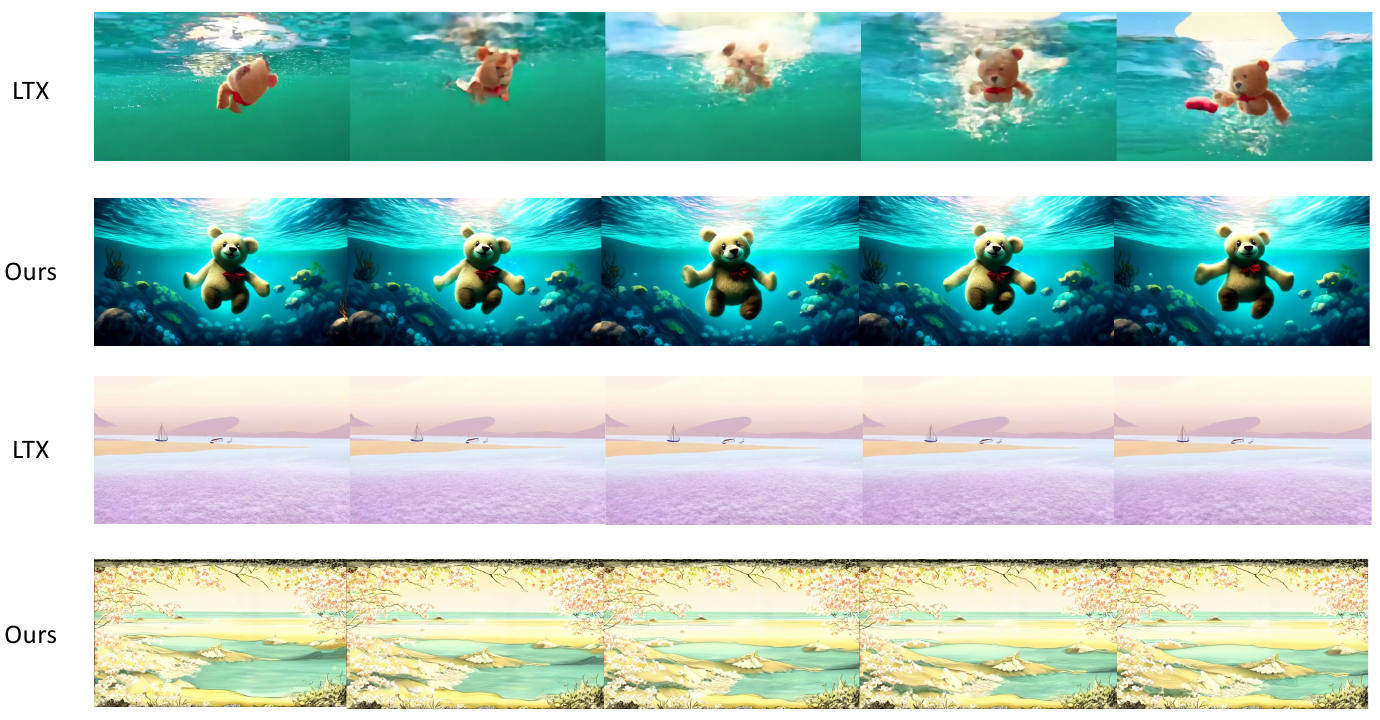}
   \caption{Qualitative comparisons between the original LTX model and our text-to-video pre-trained model. Prompt 1: A teddy bear is swimming in the ocean.; Prompt 2: A beautiful coastal beach in spring, waves lapping on sand by Hokusai, in the style of Ukiyo. Video results are attached in the supplementary material.}
   \label{fig:t2v}
\end{figure*}

As shown in Fig.~\ref{fig:t2v}, our text-to-video pre-trained model generates videos with enhanced visual fidelity and better semantic alignment, demonstrating the advantages of replacing the original text encoder and adopting our unified architecture.

We also provide detailed comparisons on the VBench~\cite{huang2023vbench} benchmark in Table~\ref{tab:eval_scores}.

\subsection{Effect of Removing Pre-training}

We conduct an ablation experiment where the model is trained directly on the NSP dataset without text-to-video pre-training. As shown in Table~\ref{tab:nsp-abla-t2v}, removing pre-training leads to a drastic performance drop, yielding a causal consistency score of only 0.06. Since our base generator is LTX, a text-to-video model, text-to-video pre-training is essential for enabling effective transfer of semantic information from the Qwen-VL understanding module to the LTX generator.

\begin{table}
\caption{Causal consistency comparison between models with and without text-to-video pre-training on the NSP test dataset.}
  \label{tab:nsp-abla-t2v}
  \centering
  \begin{tabular}{@{}lc@{}}
    \toprule
    Training Stage & Causal Consistency \\
    \midrule
    Pre-training & 0.54  \\
    SFT & 0.60 \\
    RL & 0.73 \\
    \midrule
    w/o PT & 0.06 \\
    \bottomrule
  \end{tabular}
\end{table}

\begin{table*}
\centering
\caption{Detailed Comparisons on VBench.}
\label{tab:eval_scores}
\begin{tabular}{l|c|ccc}
\toprule
\textbf{Dimension} & \textbf{LTX} & \textbf{Stage 1 } & \textbf{Stage 2} & \textbf{Stage 3} \\
\midrule
\multicolumn{5}{c}{\textit{Quality Dimensions}} \\
\midrule
Subject Consistency         & 0.898 & 0.902 & 0.960 & 0.961 \\
Background Consistency      & 0.944 & 0.957 & 0.968 & 0.970 \\
Temporal Flickering         & 0.986 & 0.986 & 0.991 & 0.993 \\
Motion Smoothness           & 0.974 & 0.987 & 0.993 & 0.993 \\
Dynamic Degree              & 0.625 & 0.675 & 0.300 & 0.303 \\
Aesthetic Quality           & 0.544 & 0.566 & 0.624 & 0.628 \\
Imaging Quality             & 0.529 & 0.427 & 0.570 & 0.573 \\
\midrule
\textbf{Quality Score}      & \textbf{0.780} & \textbf{0.782} & \textbf{0.802} & \textbf{0.805} \\
\midrule
\multicolumn{5}{c}{\textit{Semantic Dimensions}} \\
\midrule
Object Class                & 0.749 & 0.725 & 0.839 & 0.798 \\
Multiple Objects            & 0.364 & 0.152 & 0.370 & 0.412 \\
Human Action                & 0.866 & 0.692 & 0.856 & 0.872 \\
Color                       & 0.816 & 0.716 & 0.836 & 0.850 \\
Spatial Relationship        & 0.572 & 0.233 & 0.517 & 0.523 \\
Scene                       & 0.526 & 0.424 & 0.485 & 0.510 \\
Appearance Style            & 0.220 & 0.214 & 0.245 & 0.241 \\
Temporal Style              & 0.223 & 0.207 & 0.232 & 0.235 \\
Overall Consistency         & 0.246 & 0.216 & 0.245 & 0.249 \\
\midrule
\textbf{Semantic Score}     & \textbf{0.674} & \textbf{0.549} & \textbf{0.686} & \textbf{0.695} \\
\midrule
\textbf{Total Score}        & \textbf{0.759} & \textbf{0.736} & \textbf{0.779} & \textbf{0.783} \\
\bottomrule
\end{tabular}
\end{table*}

\section{Implementation Details}
We build our framework upon the Qwen-VL 2.5 (7B) model~\cite{bai2025qwen2} and the LTX Text-to-Video (0.9.0) model~\cite{hacohen2024ltx}. 
Our architecture employs 256 learnable queries and a maximum sequence length of 1024 tokens. 
For classifier-free guidance (CFG)~\cite{Ho2022ClassifierFreeDG}, 10\% of conditions are randomly dropped during training, and a CFG scale of 3.0 is applied during inference to balance generation quality and diversity. 
During inference, to ensure fair comparison with baselines, we generate 65 frames at a resolution of 832$\times$480 without applying negative guidance. The sampling step is set to 50.
Within the connector module, the RMSNorm layer weight is initialized to $\sqrt{5.5}$ and equipped with a learnable {scale factor}, initially set to 0.01. 
This design enables dynamic scaling of the connector’s output, which is essential for maintaining training stability and enhancing overall performance. 

For all the aforementioned prompts involved in reward computation, metric evaluation, and NSP dataset curation, we utilize the Claude 3.7 Sonnet model as the judge model to ensure consistent reasoning quality and alignment.

Detailed implementations for different stages are explained as follows.

\textbf{Pre-training.} For stage 1, we train the model for 3 epochs with a batch size of 32. 
For stage 2, we train for 5 epochs with a batch size of 8, and for stage 3, 2 epochs with a batch size of 8. 
The model is optimized using the Prodigy optimizer~\cite{Mishchenko2023ProdigyAE} with an initial learning rate of 1.0. 
All stages are trained on 32 80G A100 GPUs.

\textbf{Supervised Fine-tuning.} For the NSP task, we train the model on 32 80G A100 GPUs with a batch size of 8, 
using the Prodigy optimizer~\cite{Mishchenko2023ProdigyAE} with an initial learning rate of 1.0.

\textbf{Reinforcement Learning.} For reinforcement learning, the model is trained on 16 80G A100 GPUs with a gradient accumulation step of 8 for a total of 60 optimization steps. 
We use a batch size of 1, a CFG scale of 3.0, and an input resolution of $480\times832\times65$. 
The sampling step is set to 20.  
For each input, the model generates 24 video candidates using identical noise seeds. 
Among these, we adopt a Best-of-N sampling strategy ($N=8$), where the top 4 and bottom 4 samples ranked by reward scores are selected for reward optimization. 
This approach encourages the model to learn from both high- and low-quality generations, improving stability and reward sensitivity during optimization.

\section{Comparison with Related Tasks}

Compared to \textit{next shot generation}~\cite{He2025Cut2NextGN}, which predicts the immediate camera shot within the same scene focusing on short-term continuity, and \textit{multi-scene generation}~\cite{Guo2025LongCT}, which emphasizes visual consistency across different scenes, our task targets long-term causal and temporal reasoning across consecutive scenes.

\section{Efficiency Comparison}

Here, we compare the memory usage and sampling time of our model with several prior video generation models, including the original LTX model~\cite{hacohen2024ltx}, the Wan 2.1 1.3B model~\cite{Wang2025WanOA}, and the open-source unified video model Omni-Video~\cite{tan2025omni}. The results are summarized in Table~\ref{tab:efficiency}. Compared with Wan and Omni-Video, our model achieves nearly a 10× speed-up in sampling, owing to its efficient generation architecture and connector design. Relative to Omni-Video, our model also requires less GPU memory. All comparisons are conducted using an input resolution of $480\times832\times65$, 50 sampling steps, on a single 80G A100 GPU.

\begin{table}
  \caption{Efficiency comparison with previous methods.}
  \label{tab:efficiency}
  \centering
  \begin{tabular}{@{}lcc@{}}
    \toprule
    Methods & GPU Memory Usage & Time\\
    \midrule
    LTX &  14.9G & 12s \\
    Wan & 16.9G & 2m22s\\
    Omni-Video & 37.6G &2m16s \\
    Ours & 26.4G&13s \\
    \bottomrule
  \end{tabular}
\end{table}

\end{document}